\documentclass[runningheads]{llncs}

\usepackage[T1]{fontenc}
\usepackage{graphicx}
\usepackage{amsmath}
\usepackage{amssymb}
\usepackage{booktabs}
\usepackage{microtype}
\usepackage{bbding}
\graphicspath{{figures/}}

\begin{document}

\title{ResonatorLM: Causal Resonant Field Mixing for Efficient Long-Context Language Modeling}
\titlerunning{ResonatorLM}

\author{Archie Chaudhury\inst{1}\orcidID{0009-0001-1427-0664}}
\authorrunning{A. Chaudhury}
\institute{Axionic Labs\\
\email{archie@axioniclabs.ai}}
\maketitle

\begin{abstract}
Contemporary language models are dominated by the transformer architecture, which leverages self-attention mechanisms to enable more efficient, parallelized training across a wide set of documents and corpora. This has allowed transformers to effectively model data across a wide range of modalities and contexts. However, transformers, along with their conventional counterparts such as recurrent neural networks (RNNs) and convolutional neural networks (CNNs), often struggle to maintain efficiency when processing long contexts. We introduce ResonatorLM, a new mechanism that replaces attention with a physics-derived alternative. ResonatorLM treats token sequences as a single, driven one-dimensional latent field and replaces attention dot products with causal functions of damped resonators. We implement ResonatorLM on a traditional network architecture and test it on standard long-context modeling tasks. We find that in a small, 6M matched setting, training and prefill speedups increase with sequence length, decode speed reaches 6.47x compared to that of a standard, optimized transformer at 32K tokens, and accuracy reaches 61.31 percent (compared to 55.32 percent) on WikiText.
\keywords{long-context language modeling \and efficient decoding \and sequence modeling \and alternative to attention}

\end{abstract}

\section{Introduction}

Consistent, efficient long-range sequence modeling has been the primary goal of generative model architectures. Transformers have emerged as the current, most widely adopted state-of-the-art approach for modeling multimodal tasks with long contexts \cite{hoffmann2022training}. The adoption of transformers has led to the creation of generalized language models (LMs), which have shown themselves to be capable of modeling text and language across a wide range of use cases. Transformers, and by extension the majority of modern LMs, depend on the attention mechanism to efficiently parallelize training on large sets of data \cite{vaswani2017attention,brown2020language,kaplan2020scaling}. Despite the widespread adoption of transformers, their core computational profile remains inefficient and expensive at long context lengths \cite{pope2022efficiently}.

Recent alternatives have tried to achieve stronger results, mostly by modifying the attention mechanism rather than departing from it entirely. Linear and kernelized methods approximate attention with feature maps or low-rank structure \cite{katharopoulos2020transformers,choromanski2021rethinking}. Other long-context architectures use implicit convolutional filters or state-space recurrences to obtain linear-time sequence updates, including Hyena, S4, and Mamba \cite{poli2023hyena,gu2022efficiently,gu2023mamba}. These models show that explicit quadratic attention is not necessarily needed for strong sequence modeling; yet, they are still inherently derived from the existing attention-based architecture.

We introduce ResonatorLM, a language model architecture that replaces self-attention with causal resonant field mixing. ResonatorLM models a token sequence as a driven latent field and uses learned resonant kernels to transmit context across positions. This design preserves parallel computation during training and uses compact recurrent states during decoding for efficient long-context inference. This allows ResonatorLM to be significantly more efficient at language modeling, especially in long-context scenarios. Implementing ResonatorLM with a standardized language modeling architecture and benchmarking against a matched transformer, we find that it is both faster and more accurate. In a 6M-parameter WikiText-2 character setting, it reaches a decode speedup of 6.47x at 32K tokens while outperforming the baseline on both perplexity and accuracy. In a separate experiment focusing solely on a kernel reference benchmark, ResonatorLM reaches a 440.29x speedup at 8K tokens and 575.86x at 32K tokens.

\section{Related Work}

Long-context language modeling has been primarily shaped by the transformer and its algorithmic offshoots. These models leverage self-attention to give each token access to all preceding tokens within individualized layers, allowing for large-scale modeling regimes that remain efficient across various architectures and scenarios \cite{vaswani2017attention,brown2020language,kaplan2020scaling,hoffmann2022training}. Other structured attention variants include the sparse transformer, reformer, BigBird, and Longformer, as well as projection-based and kernel-based formulations such as Linformer, linear attention, and Performer \cite{child2019sparse,kitaev2020reformer,zaheer2020bigbird,beltagy2020longformer,wang2020linformer,katharopoulos2020transformers,choromanski2021rethinking}. Linear and kernelized attention replace explicit similarity matrices with feature-map or low-rank formulations.

Current alternatives to the transformer architecture often replace self-attention regimes, specifically by replacing token-to-token dot products with efficient sequence operators. Early state-space formulations established a principled route from control-theoretic structure to deep sequence modeling via HiPPO and S4 \cite{gu2020hippo,gu2022efficiently}. Structured state-space models provide linear-time sequence updates with learned recurrences: recent architectures have formalized this as state-space models (SSMs), demonstrating that recurrence and implicit convolution can remain viable for sequence modeling while reducing the compute required during both training and decoding \cite{gu2023mamba,poli2023hyena}.

Physics and continuous-time algorithms have long been emphasized in machine learning and broader sequence modeling. Neural ODEs and Hamiltonian neural networks show how differential-equation structure and conserved dynamics can improve modeling priors, stability, and interpretability in learned systems \cite{chen2018neuralode,greydanus2019hamiltonian}. ResonatorLM is directly inspired by these methods, with the motivation that embedding physical dynamics in a sequence operator can allow for both efficiency and the maintenance of computational structure.

Systems work remains essential for fair long-context claims. FlashAttention-style kernels reduce the practical overhead of exact attention on modern hardware, and decoding studies characterize the cache-driven memory bottlenecks that dominate long-context inference \cite{dao2022flashattention,dao2023flashattention2,shazeer2019fast,pope2022efficiently}. For this reason, our evaluation separates practical block benchmarks from kernel-only scaling benchmarks. This separation is critical for interpreting where gains come from: full-block wall-clock behavior versus asymptotic mixer behavior.

We implement a basic kernel-only scaling benchmark to measure pure algorithmic speedup. While this is significantly less practical than a wall-clock benchmark, it does allow us to see the algorithmic advantage of our method compared to the native baseline.\footnote{An independent proposal on the Hugging Face forums introduced a closely related wave-field sequence mixer using damped oscillatory kernels and FFT-based causal convolution, while also using physics-based diagnostics, reinforcing the broader viability of physics-derived alternatives to attention \cite{badaramoni2026wavefield}.}

\section{Model Design}

ResonatorLM replaces self-attention with a causal mixer built from damped resonant modes. The central design goal is to keep a parallel full-sequence path for training and prefill, while using fixed-size recurrent state for autoregressive decoding. Each block combines this resonant mixer with a local lexical path and a standard feed-forward sublayer.

We create a generalized model that leverages a causal resonant field mixer with per-head damped oscillatory dynamics as an alternative to self-attention. Each head is parameterized, and the resulting kernel is executed in two modes: causal FFT convolution for training and prefill, and fixed-size recurrent state updates for decoding. The same block includes a local lexical path and optional cross-head coupling, so token-level precision and long-range propagation remain in a single architecture.

At a high level, a token sequence is treated as a driven latent field rather than a set of token-to-token dot products. Learned resonant kernels transport information across positions; the local branch preserves short-range lexical detail, and optional head coupling shares information across resonant channels. The block therefore allows for two key algorithmic improvements: full-sequence training and prefill run through causal FFT convolution with $O(n \log n)$ complexity, while autoregressive decoding uses fixed-size recurrent state per head instead of an ever-growing cache. The local lexical path and cross-head coupling preserve short-range lexical detail alongside long-range propagation.

\begin{figure}[tbp]
\centering
\includegraphics[width=\textwidth]{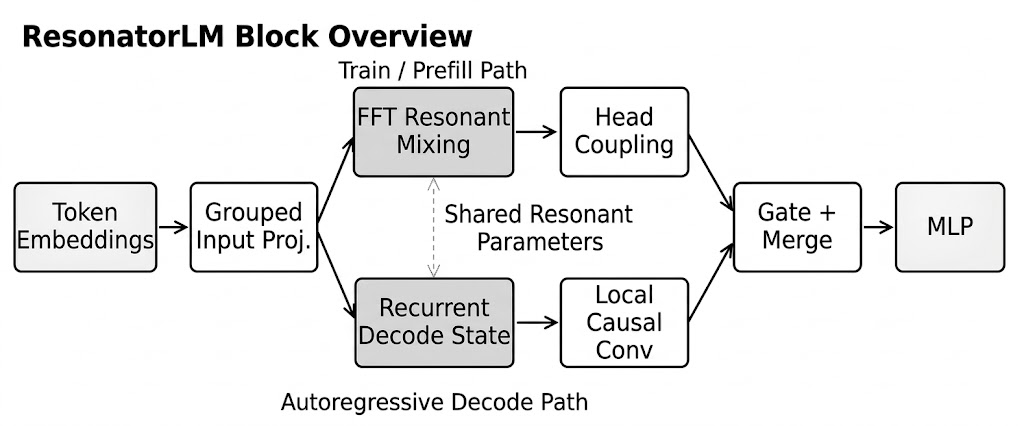}
\caption{ResonatorLM block overview. A grouped input projection feeds a resonant mixing path that uses FFT convolution during training and prefill and fixed-size recurrent state during autoregressive decoding. A local causal convolution and head coupling complement the global resonant field before the output is passed to the MLP.}
\label{fig:block_diagram}
\end{figure}

Figure~\ref{fig:block_diagram} summarizes the full block and the two execution modes used by the same kernel family: FFT-based convolution for full-sequence passes and recurrent state updates for one-step decoding.

\subsection{Resonant Kernel Parameterization}

For head $h$ and discrete lag $t \ge 0$, the mixer kernel is
\begin{equation}
k_h[t] = \exp(-\alpha_h t)\cos(\omega_h t + \phi_h),
\end{equation}
where $\alpha_h > 0$ controls damping, $\omega_h \in (0,\pi)$ is angular frequency, and $\phi_h$ is a phase offset. In implementation, unconstrained parameters are mapped to valid ranges with softplus/sigmoid transforms (with $\alpha_{\min}=10^{-4}$), then initialized with log-spaced half-lives and frequencies so different heads begin at different timescales.

The input sequence $x \in \mathbb{R}^{B \times T \times d}$ is projected by a grouped linear layer into a drive tensor $u$ and a gate tensor $g$. The gate $\gamma=\sigma(g)$ is applied after resonant mixing, which keeps a content-dependent modulation signal without constructing query-key similarity matrices.

\subsection{Full-Sequence Training and Prefill}

For training and other full-sequence passes without an existing cache, each head is mixed by causal convolution with its resonant kernel. Let $u_h \in \mathbb{R}^{T \times d_h}$ be the drive for head $h$. The mixer computes
\begin{equation}
y_h = k_h * u_h,
\end{equation}
where $*$ denotes causal convolution over the sequence dimension. The implementation evaluates this operator with FFTs:
\begin{equation}
y_h = \mathcal{F}^{-1}\!\left(\mathcal{F}(u_h)\odot\mathcal{F}(k_h)\right),
\end{equation}
using zero-padding to the next power of two. This gives an $O(T \log T)$ mixing path for training and prefill.

When the model is asked to build a cache from a prompt, it uses the same FFT path for outputs and then computes the terminal recurrent state analytically. For each head, the prompt state is the weighted sum of past drives under the complex decay factor $e^{(-\alpha_h + i\omega_h)t}$, stored as real and imaginary components. This keeps prefill and decode on the same underlying kernel family.

\subsection{Recurrent Decoding}

Autoregressive decoding uses a fixed-size per-head recurrent state instead of a growing key-value cache. For one token step, each head maintains a complex state $s_t = a_t + i b_t$ and updates it as
\begin{equation}
s_{t+1} = \lambda_h s_t + u_t,
\qquad
\hat{y}_{t+1} = \Re\!\left(e^{-i\phi_h} s_{t+1}\right),
\end{equation}
\begin{equation}
\lambda_h = \exp(-\alpha_h)\left(\cos \omega_h + i \sin \omega_h\right).
\end{equation}
In code, the state is represented as two real tensors for numerical stability and efficient batching. Its size is $B \times H \times d_h \times 2$ and does not grow with decoded length. This split between FFT convolution for full-sequence passes and recurrence for one-step decoding is central to the architecture. It preserves the same damped resonant dynamics in both regimes while avoiding the quadratic prefill cost and the linear-in-context decode memory of standard attention.
\subsection{Head Coupling and Local Lexical Path}

The resonant field alone is complemented by two additional mechanisms. First, the model can mix information across heads with a learned coupling matrix
\begin{equation}
C = I + \frac{\lambda_c}{\sqrt{H}} \tanh(\widetilde{C}),
\end{equation}
where $\lambda_c$ is the configured coupling strength. After resonant mixing, the head tensor is transformed by
\begin{equation}
\bar{y}_{g,t,:} = \sum_{h=1}^{H} C_{gh} y_{h,t,:}.
\end{equation}
Setting $\lambda_c=0$ recovers uncoupled heads. This setting appears in the ablation study.

Second, the model includes an optional causal depthwise convolution on the token stream. This local path is intended to preserve short-range lexical precision that can be weak in small global mixers. If the kernel size is $K>0$, the local branch produces $l_t$ by causal depthwise convolution, scales it channel-wise with a learned vector $s$, and adds it to the gated resonant field:
\begin{equation}
z_t = \gamma_t \odot \operatorname{flatten}(\bar{y}_t) + s \odot l_t.
\end{equation}
The combined output is then projected back to model width by a grouped output projection and passed through dropout.

At the block level, ResonatorLM uses root mean square normalization before the mixer and before the feed-forward network, residual connections around both sublayers, and a SwiGLU MLP after mixing \cite{zhang2019rmsnorm,shazeer2020glu}. The model therefore changes the sequence-mixing operator while keeping the surrounding language-model stack conventional.

\subsection{Physics Verification}

Rather than utilizing attention maps to diagnose and verify the structure of the model, our implementation leverages learned parameters from the resonant field. These parameters are listed below:
\begin{itemize}
\item impulse response per head, computed from the learned kernel;
\item head half-lives, defined as $\log 2 / \alpha_h$;
\item spectral decomposition from the FFT of the impulse response;
\item state-energy decay, given by $\exp(-2\alpha_h t)$;
\item a causality test that perturbs the suffix of an input sequence and measures the maximum change in the output prefix.
\end{itemize}
These quantities serve two roles. They verify that the mixer respects causal structure, and they provide an interpretable view of what timescales the learned modes occupy. In the reported configuration, the measured half-life range spans approximately 2 to 2048 tokens and the causality check passes with maximum prefix error below $10^{-6}$.

\section{Experimental Protocol}

We implement ResonatorLM as a matched-scale 6M-parameter language model and
compare it against a matched transformer in a study spanning quality,
robustness, breadth, and efficiency. Our primary hypothesis is that a model
using a sequence mixer derived from resonant fields can improve long-context
efficiency while maintaining quality.

Across all experimental runs, we find that ResonatorLM improves both perplexity
and accuracy. In long-context block-based benchmarks, it becomes substantially
faster, and in the kernel-tail benchmark, it shows a strong theoretical scaling speedup against the baseline.

The primary quality evaluation uses WikiText-2 raw with a character tokenizer
\cite{merity2017pointer} and a matched parameter budget near 6M. The
transformer baseline uses $d_{\mathrm{model}}=248$, 6 layers, and 8 heads.
ResonatorLM uses $d_{\mathrm{model}}=256$, 6 layers, 8 heads, the full
resonator preset, a local lexical path, and cross-head coupling. Both models are
trained for 10{,}000 steps with the same split, optimizer family, schedule, and
evaluation protocol. The main matched comparison used six seeds per model. We report
test perplexity, test top-1 accuracy, and training throughput. Parity,
transfer, long-context, and scale studies use three seeds per setting, and stress
ablations use three seeds per family.

Our pipeline remained consistent across all experiments. Each seed was run with the same configuration.
For any row in the matched, transfer, long-context, or scale studies, compared
models share train/validation/test split, tokenizer, maximum sequence length,
optimizer family, learning-rate schedule, batch size, evaluation batch size, and
number of optimization steps. The only architectural change is the sequence
mixer plus the small width adjustment needed to maintain the matched parameter
budget. The same compute settings were used for all reported results.

\begin{figure}[t]
\centering
\includegraphics[width=0.88\textwidth]{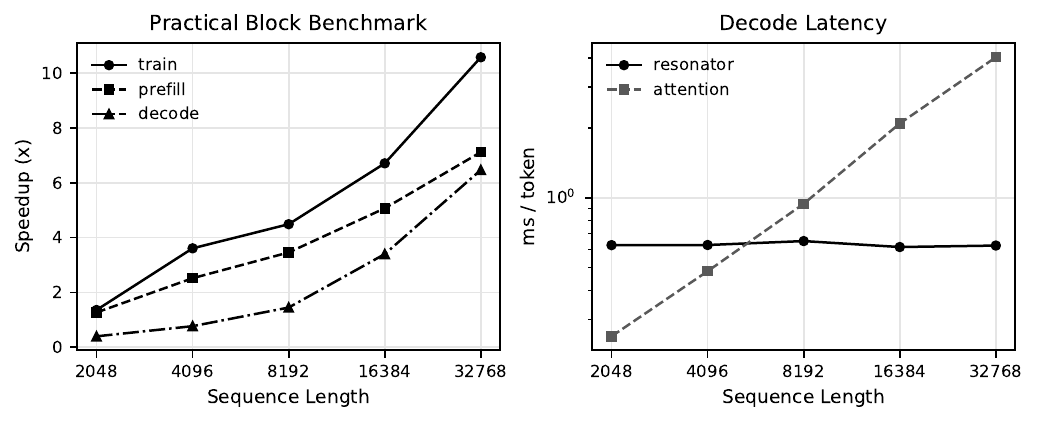}
\caption{Practical block speedup trends versus sequence length. Train and
prefill speedups increase with context length, while decode transitions from
slower at short context to faster at long context.}
\label{fig:block_speedup}
\end{figure}

We keep speed claims protocol-specific. The practical block benchmark measures
full block behavior against optimized attention, including projections,
cache/state construction, and decode-step cost. The kernel-tail benchmark
isolates asymptotic mixer scaling against a quadratic causal reference. This
separation allows us to report both pure wall-clock speedups in a practical
setting and kernel-tail results that reflect theoretical algorithmic speedup.
Figure~\ref{fig:block_speedup} provides a compact visual summary of the
practical block speedup trends used in this evaluation. Auxiliary studies extend the main setup across WikiText-103 and TinyStories \cite{merity2017pointer,eldan2023tinystories}, and test longer contexts than the 256-token matched-budget baseline.

\section{Results}

\subsection{Matched Quality and Core Architecture Tests}

Table~\ref{tab:main_quality} gives the central matched-budget result.
ResonatorLM improves both test perplexity and test accuracy over the
transformer baseline, while the transformer retains higher training throughput.
Across six seeds, the quality gap is large and stable: transformer test
perplexity has a 95\% confidence interval of $[4.561, 4.673]$ versus
$[3.757, 3.772]$ for ResonatorLM, and transformer accuracy has
$[54.968, 55.674]$ versus $[61.239, 61.390]$ for ResonatorLM.

In mean terms, ResonatorLM reduces test perplexity from 4.617 to 3.764 (about
an 18.5\% relative reduction) and increases test accuracy from 55.32\% to
61.31\% (+5.99 percentage points). The throughput tradeoff is explicit:
ResonatorLM trains at 172{,}890 tok/s versus 262{,}610 tok/s for the matched
transformer, so the quality gain is not coming from an advantage in raw training
throughput.

\begin{table}[t]
\caption{Matched 6M WikiText-2 character results. Means and standard deviations
are computed over six seeds.}
\label{tab:main_quality}
\centering
\begin{tabular}{lcccc}
\toprule
Model & Runs & Test PPL & Test Acc. (\%) & Train tok/s \\
\midrule
Transformer & 6 & $4.617 \pm 0.070$ & $55.32 \pm 0.44$ & 262610.5 \\
ResonatorLM & 6 & $3.764 \pm 0.010$ & $61.31 \pm 0.09$ & 172890.1 \\
\bottomrule
\end{tabular}
\end{table}

Table~\ref{tab:ablation_diag} tests whether this gain depends on one brittle
configuration. Across six-seed core ablations, removing coupling or the local
path changes full-model quality only modestly, and all variants remain tightly
grouped. The spread in perplexity across the three variants is only 0.021
(3.771 to 3.750), and the spread in accuracy is only 0.171 points (61.275 to
61.446). This indicates the matched-quality result is not tied to a single
fragile hyperparameter setting.

The same table reports physics diagnostics from the matched resonator checkpoint.
The measured max prefix error of $7.75\times10^{-7}$ confirms numerical
causality at strict tolerance. The learned half-life range spans 2.0 to 2048.0
tokens, showing that the model occupies both short and long timescales rather
than collapsing to a narrow temporal regime.

\begin{table}[t]
\caption{Core ablations and physics diagnostics. Left: six-seed core ablations
on WikiText-2 raw. Right: diagnostics from the matched resonator checkpoint.}
\label{tab:ablation_diag}
\centering
\begin{minipage}[t]{0.60\textwidth}
\centering
\small
\begin{tabular}{lccc}
\toprule
Variant & Runs & Test PPL & Test Acc. (\%) \\
\midrule
Balanced preset & 6 & $3.771 \pm 0.013$ & $61.275 \pm 0.094$ \\
No coupling & 6 & $3.765 \pm 0.011$ & $61.316 \pm 0.109$ \\
No local path & 6 & $3.750 \pm 0.012$ & $61.446 \pm 0.099$ \\
\bottomrule
\end{tabular}
\end{minipage}\hfill
\begin{minipage}[t]{0.35\textwidth}
\centering
\small
\begin{tabular}{lc}
\toprule
Metric & Value \\
\midrule
Max prefix error & $7.75 \times 10^{-7}$ \\
Min half-life & 2.0 \\
Max half-life & 2048.0 \\
Coupling radius & 1.0 \\
\bottomrule
\end{tabular}
\end{minipage}
\end{table}

\subsection{Breadth Across Data and Context}

The breadth studies maintain the same trend after moving to three seeds per
setting. Table~\ref{tab:breadth} shows ResonatorLM ahead across transfer
settings (WikiText-2 char/byte, WikiText-103 char, TinyStories char/byte) and
long-context settings (WikiText-2 at 512 and 1024, plus TinyStories
long-context).

In the transfer rows, the gap is consistent across datasets and tokenization
choices. On WikiText-2, ResonatorLM remains around 1.0 perplexity lower and
about 6.9--7.4 accuracy points higher under both char and byte tokenization. On
WikiText-103 char, the pattern remains similar (5.086 vs 3.856 perplexity,
52.511\% vs 60.306\% accuracy). On TinyStories, ResonatorLM keeps a clear lead
under both tokenizers, with the largest margin in byte-tokenized accuracy
(73.653\% vs 66.527\%).

In the long-context rows, the quality difference widens as tasks become harder.
At WikiText-2 length 512 and 1024, transformer perplexity rises sharply to 8.696
and 11.022, while ResonatorLM stays near 4.339 and 4.502. The same trend
appears in accuracy. The TinyStories long-context task shows the same
direction: 8.940 vs 2.950
perplexity and 33.778\% vs 66.505\% accuracy.

The shared parity sweep also preserves ordering: at learning rate
$5\times10^{-4}$ and zero dropout, the transformer reaches
$4.058 \pm 0.036$ perplexity and $59.20 \pm 0.31$\% accuracy, while
ResonatorLM reaches $3.715 \pm 0.013$ and $61.86 \pm 0.14$\%. The short-run
three-seed scale sweep keeps ResonatorLM ahead at 6M, 15M, and 30M.

\begin{table}[t]
\caption{Transfer and long-context quality. All rows report mean and standard
deviation over three seeds per setting.}
\label{tab:breadth}
\centering
\begingroup
\scriptsize
\setlength{\tabcolsep}{3pt}
\begin{tabular}{lcccc}
\toprule
Setting & Trans. PPL & Trans. Acc. & Res. PPL & Res. Acc. \\
\midrule
WikiText-2 char & $5.062 \pm 0.050$ & $52.696 \pm 0.214$ & $3.925 \pm 0.008$ & $60.116 \pm 0.048$ \\
WikiText-2 byte & $4.972 \pm 0.064$ & $53.176 \pm 0.354$ & $3.918 \pm 0.015$ & $60.084 \pm 0.091$ \\
WikiText-103 char & $5.086 \pm 0.083$ & $52.511 \pm 0.515$ & $3.856 \pm 0.009$ & $60.306 \pm 0.065$ \\
TinyStories char & $2.743 \pm 0.017$ & $68.452 \pm 0.160$ & $2.304 \pm 0.004$ & $73.566 \pm 0.040$ \\
TinyStories byte & $2.922 \pm 0.072$ & $66.527 \pm 0.699$ & $2.298 \pm 0.001$ & $73.653 \pm 0.003$ \\
\midrule
WikiText-2 512 & $8.696 \pm 0.323$ & $36.727 \pm 1.001$ & $4.339 \pm 0.010$ & $57.200 \pm 0.124$ \\
WikiText-2 1024 & $11.022 \pm 0.071$ & $29.762 \pm 0.123$ & $4.502 \pm 0.025$ & $56.166 \pm 0.182$ \\
TinyStories long-context & $8.940 \pm 0.029$ & $33.778 \pm 0.084$ & $2.950 \pm 0.013$ & $66.505 \pm 0.113$ \\
\bottomrule
\end{tabular}
\endgroup
\end{table}

\subsection{Practical Efficiency and Kernel-Tail Scaling}

Table~\ref{tab:block_speed_main} reports practical long-context block behavior.
Train and prefill speedups increase monotonically with sequence length. Decode
is slower than attention at 2048 tokens, crosses over between 4K and 8K, and then
becomes decisively faster, reaching 6.47x at 32K. This indicates that
ResonatorLM achieves significant efficiency gains specifically in long-context
modeling scenarios.

The decode crossover is visible directly in ms/token values. At a context size of 2048 tokens,
ResonatorLM decode is 0.6254 ms/tok versus 0.2532 for attention. At 8192, the
ordering flips (0.6511 vs 0.9423), and the advantage grows at 16384
(0.6138 vs 2.0897) and 32768 (0.6228 vs 4.0296). The prefill and train columns
show the same context-length trend: practical gains strengthen as sequence
length increases. All speed tables were measured on a single NVIDIA L4 GPU using PyTorch
2.10.0+cu128. Training uses CUDA automatic mixed precision with bfloat16. The
kernel-tail benchmark uses float32 and reports the winning chunk size by
sequence length in Table~\ref{tab:kernel_tail_main}. The practical block
benchmark uses batch size 1, decode length 128, five timed iterations after two
warm-up iterations, and synchronized wall-clock timing with explicit device
synchronization around timed regions. The baseline uses PyTorch
scaled-dot-product attention, while ResonatorLM uses cuFFT-backed
\texttt{torch.fft} for full-sequence passes and fixed-size recurrent state for
decode.

\begin{table}[t]
\caption{Practical block benchmark against an optimized attention block. This
protocol includes projections, state or cache construction, and decode cost.}
\label{tab:block_speed_main}
\centering
\small
\begin{tabular}{rrrrrr}
\toprule
Seq. Len. & Train & Prefill & Decode & Res. ms/tok & Attn. ms/tok \\
\midrule
2048 & 1.36x & 1.27x & 0.40x & 0.6254 & 0.2532 \\
4096 & 3.61x & 2.52x & 0.77x & 0.6262 & 0.4829 \\
8192 & 4.49x & 3.46x & 1.45x & 0.6511 & 0.9423 \\
16384 & 6.71x & 5.07x & 3.40x & 0.6138 & 2.0897 \\
32768 & 10.58x & 7.12x & 6.47x & 0.6228 & 4.0296 \\
\bottomrule
\end{tabular}
\end{table}

We implement a separate kernel-tail benchmark to measure pure long-context
speedup against a baseline asymptotic mixer, without overhead from projections
or multiple layers. Table~\ref{tab:kernel_tail_main} shows that under the
disclosed reference setting (float32, chunk size 32), the resonator kernel
reaches 440.29x at 8K tokens and 575.86x at 32K tokens.

\begin{table}[t]
\caption{Kernel-tail benchmark against a quadratic causal reference. This
protocol is separate from the practical block benchmark and should not be
interpreted as an end-to-end model speedup.}
\label{tab:kernel_tail_main}
\centering
\begin{tabular}{rccc}
\toprule
Seq. Len. & Dtype & Chunk & Best Speedup \\
\midrule
8192 & torch.float32 & 32 & 440.29x \\
32768 & torch.float32 & 32 & 575.86x \\
\bottomrule
\end{tabular}
\end{table}

\section{Discussion}

Our results show that a non-attention mixer with a physics-motivated foundation
can outperform a transformer on language quality while also improving
long-context efficiency. Across multiple runs, ResonatorLM consistently
outperforms or matches a transformer baseline and shows a clear decode-speed
advantage in long contexts. To test whether these gains come from the resonant
mixer rather than auxiliary architectural choices, we conducted a multi-step
ablation study that removed coupling and the local path; these changes produced
only modest performance shifts. We also evaluated both practical long-context
regimes and a separate kernel benchmark, showing that the method remains strong
across settings and can realize much larger asymptotic speedups.

\paragraph{Limitations} This work is intended as a foundational demonstration
that a physics-derived alternative to attention can remain competitive in
language modeling while improving long-context algorithmic efficiency. Despite
statistically significant results across multiple seeds, our experiments cover a
limited benchmark set and small model sizes. Future work should test
ResonatorLM at larger scales and on more complex long-context tasks, as well as
on broader sequence-modeling domains, including multimodal settings. We also
compare primarily against a matched transformer baseline; evaluating against
newer architectures and stronger system-level optimizations is an important
next step. Furthermore, we note that the practical block benchmark is not an actual production benchmark, where other factors may impact the actual speedup that is observed. As such, our work is meant to be foundational in nature, rather than serving as a clear empirical proof that Resonator has significant advantages over a transformer or other similar architectures.

\section{Conclusion}

In this work, we implemented ResonatorLM, a language model architecture that
replaces attention with causal resonant field mixing while preserving
consistent dynamics across FFT-based sequence modeling and recurrent decoding.
We first formalized ResonatorLM and introduced our design. We then benchmarked
it across multiple language-modeling tasks, including WikiText-2,
WikiText-103, and TinyStories. Across three-seed breadth studies, the same
trend holds under tokenizer changes, longer contexts, and short-run scaling. In
the practical block benchmark, long-context efficiency grows with sequence
length and reaches a 6.47x decode speedup at 32K. In the separate kernel-tail
benchmark, the resonator kernel substantially exceeds reported long-context
targets.

\begin{credits}
\subsubsection{\discintname}
The authors have no competing interests to declare that are relevant to the content of this article.
\subsubsection{Usage of LLMs}
We utilized AI assistants to help draft, proofread, and formalize some of the key figures in our paper. All core ideas were implemented by the authors.
\end{credits}

\bibliographystyle{splncs04}
\bibliography{resonator_refs}

@inproceedings{vaswani2017attention,
  author    = {Ashish Vaswani and Noam Shazeer and Niki Parmar and Jakob Uszkoreit and Llion Jones and Aidan N. Gomez and Lukasz Kaiser and Illia Polosukhin},
  title     = {Attention Is All You Need},
  booktitle = {Advances in Neural Information Processing Systems},
  year      = {2017}
}

@inproceedings{brown2020language,
  author    = {Tom B. Brown and Benjamin Mann and Nick Ryder and Melanie Subbiah and Jared Kaplan and Prafulla Dhariwal and Arvind Neelakantan and Pranav Shyam and Girish Sastry and Amanda Askell and others},
  title     = {Language Models are Few-Shot Learners},
  booktitle = {Advances in Neural Information Processing Systems},
  year      = {2020}
}

@article{kaplan2020scaling,
  author  = {Jared Kaplan and Sam McCandlish and Tom Henighan and Tom B. Brown and Benjamin Chess and Rewon Child and Scott Gray and Alec Radford and Jeffrey Wu and Dario Amodei},
  title   = {Scaling Laws for Neural Language Models},
  journal = {arXiv preprint arXiv:2001.08361},
  year    = {2020}
}

@article{hoffmann2022training,
  author  = {Jordan Hoffmann and Sebastian Borgeaud and Arthur Mensch and Elena Buchatskaya and Trevor Cai and Eliza Rutherford and Diego de Las Casas and Lisa Anne Hendricks and Johannes Welbl and Aidan Clark and others},
  title   = {Training Compute-Optimal Large Language Models},
  journal = {arXiv preprint arXiv:2203.15556},
  year    = {2022}
}

@inproceedings{dao2022flashattention,
  author    = {Tri Dao and Daniel Y. Fu and Stefano Ermon and Atri Rudra and Christopher Re},
  title     = {FlashAttention: Fast and Memory-Efficient Exact Attention with IO-Awareness},
  booktitle = {Advances in Neural Information Processing Systems},
  year      = {2022}
}

@article{shazeer2019fast,
  author  = {Noam Shazeer},
  title   = {Fast Transformer Decoding: One Write-Head is All You Need},
  journal = {arXiv preprint arXiv:1911.02150},
  year    = {2019}
}

@article{pope2022efficiently,
  author  = {Reiner Pope and Sholto Douglas and Aidan Craik and Ivan Grisly and Matthew Hall and Francesco Salimbeni and Yuhuai Wu},
  title   = {Efficiently Scaling Transformer Inference},
  journal = {arXiv preprint arXiv:2211.05102},
  year    = {2022}
}

@inproceedings{katharopoulos2020transformers,
  author    = {Angelos Katharopoulos and Apoorv Vyas and Nikolaos Pappas and Francois Fleuret},
  title     = {Transformers are RNNs: Fast Autoregressive Transformers with Linear Attention},
  booktitle = {International Conference on Machine Learning},
  year      = {2020}
}

@inproceedings{choromanski2021rethinking,
  author    = {Krzysztof Choromanski and Valerii Likhosherstov and David Dohan and Xingyou Song and Andreea Gane and Tamas Sarlos and Peter Hawkins and Jared Davis and Afroz Mohiuddin and Lukasz Kaiser and others},
  title     = {Rethinking Attention with Performers},
  booktitle = {International Conference on Learning Representations},
  year      = {2021}
}

@inproceedings{gu2022efficiently,
  author    = {Albert Gu and Karan Goel and Christopher Re},
  title     = {Efficiently Modeling Long Sequences with Structured State Spaces},
  booktitle = {International Conference on Learning Representations},
  year      = {2022}
}

@article{gu2023mamba,
  author  = {Albert Gu and Tri Dao},
  title   = {Mamba: Linear-Time Sequence Modeling with Selective State Spaces},
  journal = {arXiv preprint arXiv:2312.00752},
  year    = {2023}
}

@inproceedings{poli2023hyena,
  author    = {Michael Poli and Stefano Massaroli and Eric Q. Nguyen and Daniel Y. Fu and Tri Dao and Stephen Baccus and Yoshua Bengio and Stefano Ermon and Christopher Re},
  title     = {Hyena Hierarchy: Towards Larger Convolutional Language Models},
  booktitle = {International Conference on Machine Learning},
  year      = {2023}
}

@inproceedings{merity2017pointer,
  author    = {Stephen Merity and Caiming Xiong and James Bradbury and Richard Socher},
  title     = {Pointer Sentinel Mixture Models},
  booktitle = {International Conference on Learning Representations},
  year      = {2017}
}

@article{eldan2023tinystories,
  author  = {Ronen Eldan and Yuanzhi Li},
  title   = {TinyStories: How Small Can Language Models Be and Still Speak Coherent English?},
  journal = {arXiv preprint arXiv:2305.07759},
  year    = {2023}
}

@article{zhang2019rmsnorm,
  author  = {Biao Zhang and Rico Sennrich},
  title   = {Root Mean Square Layer Normalization},
  journal = {Advances in Neural Information Processing Systems},
  year    = {2019}
}

@article{shazeer2020glu,
  author  = {Noam Shazeer},
  title   = {GLU Variants Improve Transformer},
  journal = {arXiv preprint arXiv:2002.05202},
  year    = {2020}
}

@article{beltagy2020longformer,
  author  = {Iz Beltagy and Matthew E. Peters and Arman Cohan},
  title   = {Longformer: The Long-Document Transformer},
  journal = {arXiv preprint arXiv:2004.05150},
  year    = {2020}
}

@article{zaheer2020bigbird,
  author  = {Manzil Zaheer and Guru Guruganesh and Avinava Dubey and Joshua Ainslie and Chris Alberti and Santiago Ontanon and Philip Pham and Anirudh Ravula and Qifan Wang and Li Yang and Amr Ahmed},
  title   = {Big Bird: Transformers for Longer Sequences},
  journal = {arXiv preprint arXiv:2007.14062},
  year    = {2020}
}

@article{dao2023flashattention2,
  author  = {Tri Dao},
  title   = {FlashAttention-2: Faster Attention with Better Parallelism and Work Partitioning},
  journal = {arXiv preprint arXiv:2307.08691},
  year    = {2023}
}

@article{child2019sparse,
  author  = {Rewon Child and Scott Gray and Alec Radford and Ilya Sutskever},
  title   = {Generating Long Sequences with Sparse Transformers},
  journal = {arXiv preprint arXiv:1904.10509},
  year    = {2019}
}

@inproceedings{kitaev2020reformer,
  author    = {Nikita Kitaev and Lukasz Kaiser and Anselm Levskaya},
  title     = {Reformer: The Efficient Transformer},
  booktitle = {International Conference on Learning Representations},
  year      = {2020}
}

@article{wang2020linformer,
  author  = {Sinong Wang and Belinda Z. Li and Madian Khabsa and Han Fang and Hao Ma},
  title   = {Linformer: Self-Attention with Linear Complexity},
  journal = {arXiv preprint arXiv:2006.04768},
  year    = {2020}
}

@article{gu2020hippo,
  author  = {Albert Gu and Tri Dao and Stefano Ermon and Atri Rudra and Christopher Re},
  title   = {HiPPO: Recurrent Memory with Optimal Polynomial Projections},
  journal = {Advances in Neural Information Processing Systems},
  year    = {2020}
}

@inproceedings{chen2018neuralode,
  author    = {Ricky T. Q. Chen and Yulia Rubanova and Jesse Bettencourt and David Duvenaud},
  title     = {Neural Ordinary Differential Equations},
  booktitle = {Advances in Neural Information Processing Systems},
  year      = {2018}
}

@inproceedings{greydanus2019hamiltonian,
  author    = {Sam Greydanus and Misko Dzamba and Jason Yosinski},
  title     = {Hamiltonian Neural Networks},
  booktitle = {Advances in Neural Information Processing Systems},
  year      = {2019}
}

@misc{badaramoni2026wavefield,
  author       = {Avinash Badaramoni},
  title        = {Wave Field LLM --- O(n log n) attention via wave equation dynamics, within 5\% of standard transformer},
  howpublished = {\url{https://discuss.huggingface.co/t/wave-field-llm-o-n-log-n-attention-via-wave-equation-dynamics-within-5-of-standard-transformer/173625}},
  note         = {Hugging Face Forums post, accessed 2026-03-31},
  year         = {2026}
}

\end{document}